\documentclass[letterpaper, 10 pt, journal, twoside]{IEEEtran}
\usepackage{gensymb}
\usepackage{cclicenses, graphicx}
\usepackage{amsmath}
\usepackage{array}
\usepackage{flushend}
\usepackage{capt-of}
\usepackage{tabularx}
\usepackage{multirow}
\usepackage[caption=false]{subfig}
\usepackage{balance}
\usepackage{color}  

\usepackage[textsize=scriptsize, bordercolor=white,backgroundcolor=gray!30,linecolor=black,colorinlistoftodos]{todonotes}  
\usepackage[normalem]{ulem} 
\usepackage{soul} 

\usepackage{xcolor}

 \newcommand{\remind}[2]{{{#2}}}
 \newcommand{\corrlab}[2]{{{#2}}}
 
 \newcommand{\cusst}[1]{{}}

\newenvironment{mylist}
{\begin{list}{$\bullet$}{\labelsep2.3mm \leftmargin4mm \itemsep0.5mm \itemindent-1.7mm \topsep0mm \parsep0mm}}
{\end{list}}

\newcolumntype{C}[1]{>{\centering\let\newline\\\arraybackslash\hspace{0pt}}m{#1}}

\begin{document}
\title{Multimodal Sensing and Interaction for a Robotic Hand Orthosis}

\author{Sangwoo Park$^{1}$, Cassie Meeker$^{1}$, Lynne M. Weber$^{2}$, Lauri Bishop$^{2}$, Joel Stein$^{2,3}$, and Matei Ciocarlie$^{1,3}$%
\thanks{Manuscript received: July, 31, 2018; Revised November, 3, 2018; Accepted December, 2, 2018.}
\thanks{This paper was recommended for publication by Editor Allison M. Okamura upon evaluation of the Associate Editor and Reviewers' comments.
This work was supported in part by the National Science Foundation under grant IIS-1526960 (part of the National Robotics Initiative).}%
\thanks{$^{1}$Department of Mechanical Engineering, Columbia University, New York, NY 10027, USA.}%
\thanks{\hspace{-3mm}{\tt\footnotesize \{sp3287, cgm2144, matei.ciocarlie\}@columbia.edu}}%
\thanks{$^{2}$Department of Rehabilitation and Regenerative Medicine, Columbia University, New York, NY 10032, USA. {\tt\footnotesize \{lw2739, lb2413, js1165\}@cumc.columbia.edu}}%
\thanks{$^{3}$Co-Principal Investigators}
\thanks{Digital Object Identifier (DOI): see top of this page.}
}

\maketitle

\begin{abstract}
Wearable robotic hand rehabilitation devices can allow greater freedom
and flexibility than their workstation-like counterparts. However, the
field is generally lacking effective methods by which the user can
operate the device: such controls must be effective, intuitive, and
robust to the wide range of possible impairment patterns. Even when
focusing on a specific condition, such as stroke, the variety of
encountered upper limb impairment patterns means that a single sensing
modality, such as electromyography (EMG), might not be sufficient to enable
controls for a broad range of users. To address this significant gap,
we introduce a multimodal sensing and interaction paradigm for an
active hand orthosis. In our proof-of-concept implementation, EMG is
complemented by other sensing modalities, such as finger bend and
contact pressure sensors. We propose multimodal interaction methods
that utilize this sensory data as input, and show they can enable
tasks for stroke survivors who exhibit different impairment
patterns. We believe that robotic hand orthoses developed as
multimodal sensory platforms with help address some of the key
challenges in physical interaction with the user.

\end{abstract}

\begin{IEEEkeywords}
	Wearable Robots, Prosthetics and Exoskeletons, Rehabilitation Robotics
\end{IEEEkeywords}

\section{Introduction}

\IEEEPARstart{R}{obotic} devices for hand rehabilitation promise to alleviate some of
the critical challenges of traditional rehabilitation paradigms. In
particular, they could significantly increase the number of training
exercises for cases where access to a therapist is limited. Recent
advances~\cite{park2016, biggar2016, in2015, kang2016, zhao2016,
  polygerinos2015} have greatly improved the wearability of such
orthoses: we now have devices that provide the needed actuation
capabilities in a compact, wearable package, allowing greater freedom
and flexibility than their workstation-like counterparts. Such wearable
devices could allow use beyond the confines of a therapist's office.

However, the vision of a wearable orthotic device used for activities
of daily living (ADLs) can only be realized if the patients are able
to operate the device themselves. Control methods must be effective
and intuitive, robust to long term operation, and cannot impose
significant cognitive load. These algorithms must also cope with a
wide range of impairment levels and abilities in the target
population. While the actuation abilities of robotic hand orthoses
have made great strides, control algorithms have not made similar
progress in addressing these challenges.

The key to intuitive, user-driven control for a wearable orthosis lies
in the ability to \textit{infer the user's intent} from sensor data
collected by the device. The robotic orthosis thus becomes a
\textit{sensory platform} in addition to an actuation mechanism. The
control algorithm must infer the intent of the user from the collected
data and respond with the appropriate actuation commands. In our own
previous work, we developed a tendon-driven hand
orthosis~\cite{park2016} that used a single sensing modality, forearm
surface electromyography (EMG), to infer user
intent~\cite{meeker2017}. Other studies (which we review in the next
section) have also investigated EMG for control of wearable robotic
devices.

While this body of work has shown that intuitive control is indeed
possible, it has also highlighted numerous challenges. For example,
EMG signals are inherently abnormal in hemiparesis and distorted by
spasticity and fatigue\corrlab{R2-1}{~\cite{cesqui2013},
  ~\cite{ochoa2009}}. If signal patterns drift or change between
training and deployment, the control method has no way of coping
without new calibration or training data. Physical interaction with
the orthosis also alters the signals. In fact, other unimodal interaction methods face similar challenges : if the nature of the impairment
(which varies greatly between individuals) is such that the signal
exhibits too much or too little variation, the entire device can
become unusable.

\begin{figure}[t]
	\centering
	\begin{tabular}{c}
		\includegraphics[width=1.0\linewidth]{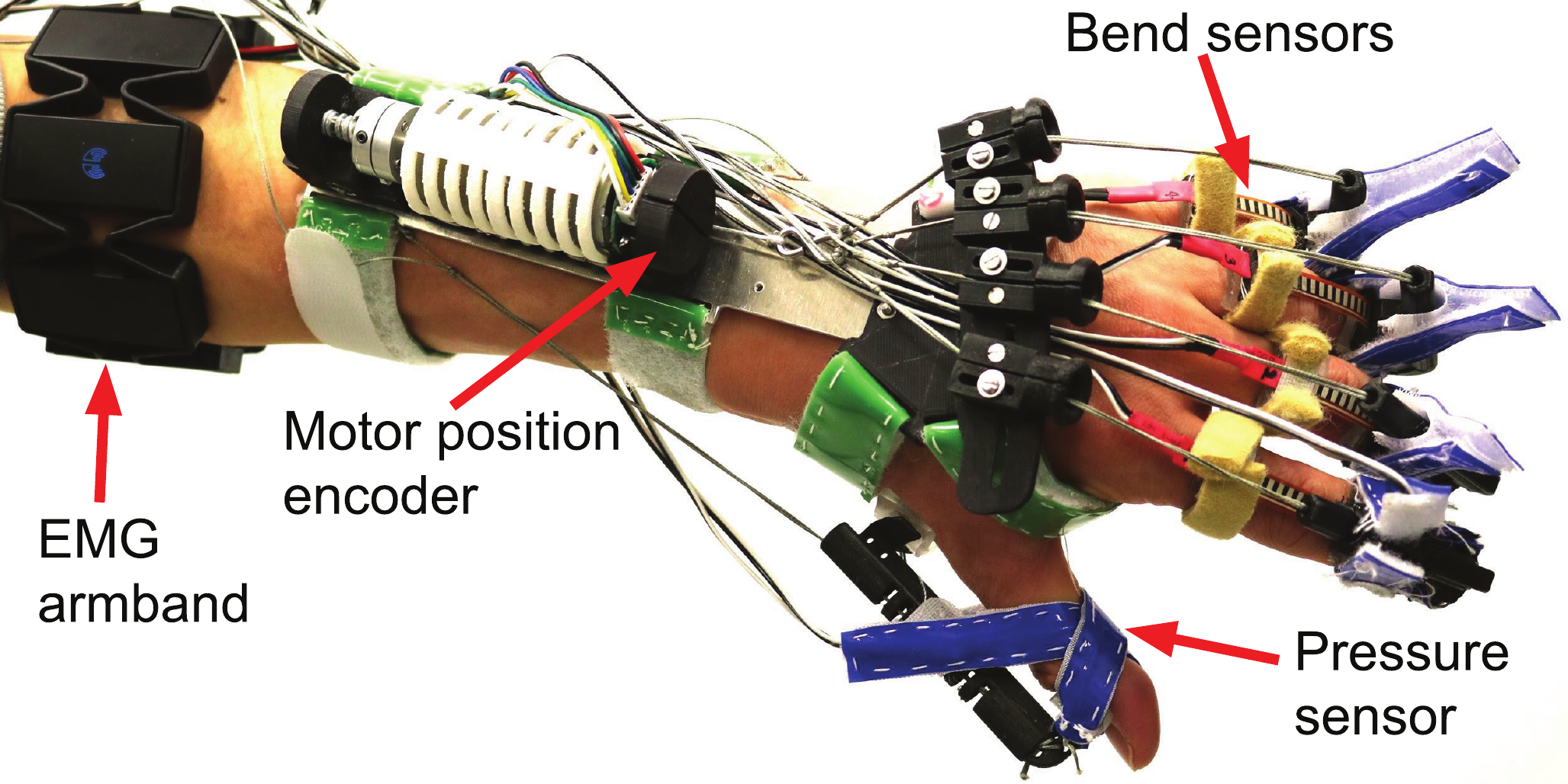}
	\end{tabular}
	\caption{Prototype of a hand orthotic device with multimodal sensors.}
	\label{fig:sensor_implementations}
\vspace{-0.6cm}
\end{figure}

\corrlab{R1-1b}{We were thus motivated to research and develop various
  forms of sensing on a wearable hand orthosis for intent detection,
  where different sensing modalities can complement and augment each
  other. We believe that this multimodal sensing approach} can help
address the aforementioned key challenges for robust, intuitive
user-driven operation. In this study, we aim for test of feasibility
by introducing a multisensory implementation developed for stroke
patients.

While stroke subjects display a wide variety of impairment patterns,
we have observed that many retain subtle, but consistent residual
movements (e.g. partial extension of one or two fingers) or patterns
of co-contraction that typically appear when a subject is prompted to
open or close the hand. To measure these abilities, we outfit an
exotendon device with bend and pressure sensors. When EMG is
insufficient to determine user intention, bend and pressure sensors
can be used to control the orthosis. We refer to controls that use
multiple sensor types for input as multimodal controls. The main
contributions of this paper are:
\begin{mylist}
\item We develop an active hand orthosis as a multimodal sensory
  platform as well as an actuation device, allowing us to characterize
  physical interaction with the user in novel ways. In particular, we
  incorporate bend and pressure sensors into an exotendon framework
  with existing EMG sensing, while keeping the orthosis compact and without impacting grasping tasks.
\item We introduce multimodal control methods for the orthosis, using
  the various sensors (EMG, bend, and pressure) as inputs. We then
  show that the different controls can be used with different
  impairment patterns commonly found in stroke subjects.
\end{mylist}
\corrlab{R2-1}{To the best of our knowledge, we are the first to
  propose intuitive multimodal control schemes for a hand orthosis
  which leverage natural hand movement signals (as opposed to side
  channels such as voice). A very recent review of more than 80
  studies in this area~\cite{chu2018} found a single device capable of
  multimodal intent inferral, and that was using voice as a second
  modality. We thus aim to bridge this gap towards reliable and
  intuitive control. Working with stroke patients, we show that our
  methods can be adapted to various impairment patterns, and can
  also be integrated in fully functional systems, laying the
  foundation for further development in this direction.}

\section{Related Work}

EMG is one of the most popular unimodal controls for robotic hand
orthoses, as it requires relatively simple algorithms and enables
intuitive operation. Most commonly, sensors are attached to the flexor
and extensor muscles of the impaired arm and an open-loop control
opens and closes the hand when EMG exceeds a
threshold~\cite{polygerinos2015, hu2013}. Pattern recognition
algorithms are also becoming more popular as they can enable the use
of commodity EMG armbands~\cite{meeker2017} and classify multiple hand
postures in stroke patients~\cite{lu2018}.

However, these algorithms often only work on a subset of stroke
population due to abnormal muscle activation~\cite{ochoa2009}. Several
strategies have been developed to adapt to these irregular EMG
patterns. One strategy is to place the sensors on muscles which retain
healthy EMG patterns. For example, stroke subjects can utilize the
contralateral upper extremity~\cite{lucas2004} or facial
expressions~\cite{hussain2017} to trigger EMG-based controls.  Both of
these methods require learning a control which uses muscles unrelated
to the desired task.

An alternative strategy is to develop a multimodal control that uses EMG in addition to 
a more robust sensing modality. The VAEDA glove uses voice recognition 
to specify the control mode, and EMG signals to trigger
commands~\cite{thielbar2017}. Voice recognition is robust in ideal
conditions, but sensitive to noise. Radio frequency identification (RFID) tags on objects
can serve as non-biological switches to identify desired hand
postures, again using EMG as a trigger to execute these
postures~\cite{yap2016a}. RFID tags predetermine which objects the
subject can interact with, which limits their utility in real-world
environments. Fusing mechanomyography (MMG) and EMG for prosthetic controls has 
been studied~\cite{guo2017, wolczowski2017}. MMG is more robust to 
noise than EMG, but its use for individuals with neurological impairment is largely unexplored\corrlab{R1-2a}{~\cite{ibitoye2014}}.

Other studies have developed controls which rely on types of
sensors other than EMG. Some of these controls are unimodal - they trigger
the device using a simple analog button~\cite{kang2016}, a bend sensor
on the wrist~\cite{in2015}, body-powered motions~\cite{luo2005}, or
force myography~\cite{yap2016b}. The Soft Extra Muscle Glove uses
force sensitive resistors (FSRs) as a control because they provide
useful information when subjects interact with
objects~\cite{nilsson2012}. Zhao et al.~\cite{zhao2016} integrated
optical strain sensors into a rehabilitation device based on pneumatic
actuation in order to provide position feedback for control and motion
analysis. However, unimodal controls have not yet been shown to be
robust for long-term operation, and often rely on external cues, 
instead of natural hand motions.

Other works have developed multimodal controls using non-EMG
sensors. For example, Steinkamp, et al.~\cite{steinkamp2014} 
use a 3D depth-camera and IMU sensors to analyze point
clouds of the environment in order to classify appropriate hand
assistance. Some devices use sensors not as control inputs,
but as tools to analyze hand movement. The SCRIPT passive orthosis is
equipped with multimodal sensors to estimate joint rotations and
torques. These sensing capabilities enable interactive rehabilitation
games for users~\cite{ates2017, amirabdollahian2014}.

\corrlab{R2-9}{Multimodal controls for gait assistive devices are more
  commonplace than for hand devices. Hybrid Assistive Limb utilizes
  pressure sensors and potentiometers to measure joint angles for
  motion intent estimation~\cite{suzuki2007}. Villa-Parra, et al. have
  developed a knee device for gait rehabilitation using EEG and EMG as
  a multimodal control~\cite{villa2015}. While multimodal intent
  inferral methods for lower-limb exoskeletons exist, hand devices
  face significantly different challenges, such as many more
  articulation degrees of freedom, wider variety of movement patterns,
  more limited space and acceptable weight, etc. This perhaps helps
  explain the fact that no similar multimodal controls have been
  introduced to date for assistive hand devices~\cite{chu2018}.}

\section{Exotendon Device} \label{exotendon}
To equip a hand orthosis with multimodal sensing, we expand upon our previous work with exotendon hand devices, specifically our work on tendon networks~\cite{park2016} combined with distal structures for efficient force transmission~\cite{park2018}. \corrlab{R1-1a}{Tendon-driven systems require less space than linkage-based exoskeletons, as they utilize few, small anchoring structures. Therefore, they are well-suited for sensor implementation.}

\begin{figure}[t]
	\centering
	\begin{tabular}{c}
		\includegraphics[width=1.0\linewidth]{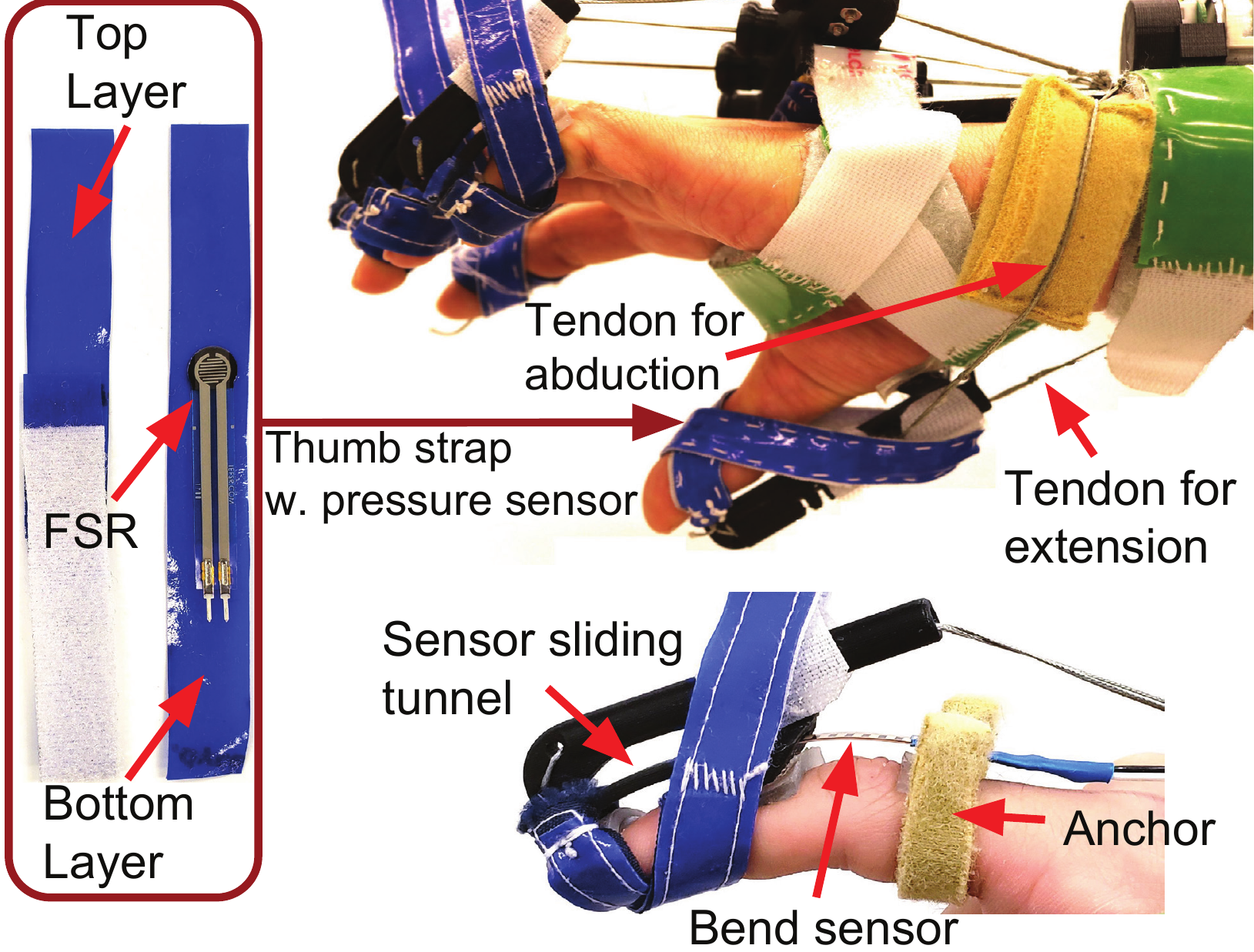}\vspace{-0.1cm}\\
	\end{tabular}
	\caption{\corrlab{R1-3a}{Integration of a pressure sensor into a thumb strap (left), tendon routing for thumb extension and abduction (top)} and bend sensor implementation (bottom).}
	\vspace{-0.5cm}
	\label{fig:sensors}
\end{figure}

We developed a modularized device consisting of two parts: an aluminum
forearm splint with actuation and 3D-printed fingertip components for
cable routing (Fig.~\ref{fig:sensor_implementations},~\ref{fig:sensors}). The splint constrains wrist movement so
that the motor forces are transmitted to the
fingers. \remind{R1-4b-ii}{For actuation, we use a
  Proportional-Integral-Derivative (PID) position controller whose
  range of motion is determined by user hand size.}
\corrlab{R1-4b-i}{Motor extension or retraction takes
  approximately 1.8 seconds.}

The 3D-printed fingertip components are secured to the fingertips
using Velcro straps. The underside of the strap is rubber to prevent
distal migration. The components route the exotendons through raised
pathways that enhance force transmission by increasing the
moment arm around the proximal interphalangeal (PIP) joints. In
addition, the components prevent hyper-extension of the distal
interphalangeal (DIP) joint and serve as an anchoring point for the
tendons.

The thumb moves differently than the other four fingers, and therefore 
requires different routing. As long as the four fingers are sufficiently 
extended, we can enable grasping tasks by simply splinting the thumb in 
a stationary, opposed position~\cite{arata2013}. We splint the thumb using 
two tendon routes which adjust the thumb's abduction and extension 
(Fig.~\ref{fig:sensors}).

\section{Multimodal Sensing} \label{multimodal}


While existing work has focused primarily on robotic hand
orthoses as actuation devices, we envision future devices serving an
equally important role as sensory platforms, equipped to characterize
physical interaction with the user. Numerous sensing modalities can be
envisioned, focusing on tendons, joints, contacts, etc. In this
context, we have developed a multisensory platform prototype,
combining sensors for the following: forearm EMG, motor position,
fingertip pressure, and joint angles
(Fig.~\ref{fig:sensor_implementations},~\ref{fig:sensors}). We describe these
sensing modalities and their integration with the orthosis next.

\subsubsection{Forearm EMG}
EMG is one of the most common orthotic controls because it is
intuitive. EMG sensors are low profile, and commercial devices, like
the one used in this work, are easy to don and doff. With relatively
simple algorithms, EMG sensors can be used to identify a variety of
different hand poses.

We use the Myo Armband from Thalmic labs for our EMG sensing. The
armband consists of eight EMG sensors and is placed on the subject's
forearm, proximal to the splint. Our pattern recognition algorithm
(Section~\ref{subsec:emg_control}) uses the EMG sensors to predict the
user's intended hand state. Fig.~\ref{fig:emg_training} shows an
example of EMG activation patterns as a subject attempts to open and
close their affected hand. In this figure, the EMG activations for open
and close are distinct; however, these patterns will change over time
as the subject fatigues.

\subsubsection{Motor Position}
Motor position sensing is commonplace in robotic devices, and we
include its description here for completeness. The motor encoder
provides high-resolution position feedback, which enables us to
control the actuator with position control and determine the current
state of the orthosis. Because our tendon network is underactuated,
this feedback does not provide information about individual finger
behaviors, but their combined movement pattern.

\subsubsection{Finger Joint Angles}
Joint angles can serve as cues to determine patient intent. One
typical pattern is partial voluntary movement, where patients try to
open their hand and some fingers partially extend. Another, abnormal,
movement pattern from which the sensing modality can potentially
benefit is overactive stretch response~\cite{kamper2006}, which
exhibits finger flexion when patients try to extend. By
measuring PIP joint angles with bend sensors, both movement patterns
can give us information about user intent.

We use a bend-sensitive resistor on each finger to measure joint flexion of the PIP joint (Fig.~\ref{fig:sensors}). We assume residual movement of the PIP joint is greater than the MCP joint and therefore only deploy sensors on the PIP. For each finger, the proximal side of the sensor is anchored to
the subject's proximal finger link by a strap. The distal side of the
bend sensor is fed through a flat hole in the bottom of the fingertip
component to keep it close to the distal link of the finger.

We found that using a simple threshold on the raw bend sensor data to
trigger an open command was limited as a control 
because motor position and the size of the objects with which the user
interacts both dramatically affect the raw data
values. Fig.~\ref{fig:bend_training} shows the raw bend data and
bend derivative during an example open-close motion. Note that the
bend derivative peaks soon after the subject is asked to open. The
next notable maximum is caused by the device extending the fingers.

\subsubsection{Fingertip Pressure}
Pressure sensors on the fingertips serve a dual role: since the digit
straps are the conduit by which exotendons apply force to the fingers,
the pressure sensor can record the level of force between the hand and
the device. When the user is performing a grasp, the pressure sensors
will also record the contact force between the hand and the object. In
this way, pressure sensing allows us to paint a complete picture of
force transmission, from the orthosis to the patient's hand, and from the
hand to the environment.

Fingertip pressure increases when the subject is either
interacting with an object or trying to close the hand while the device is open. Though we
cannot differentiate between the two actions, the increase in pressure
gives us useful information about when the user intends to close their
hand, especially when the user cannot maintain the muscle activation
necessary for detection via EMG.

\begin{figure}[t]
	\centering
	\subfloat[Forearm EMG activation]{%
		\includegraphics[width=1.0\linewidth]{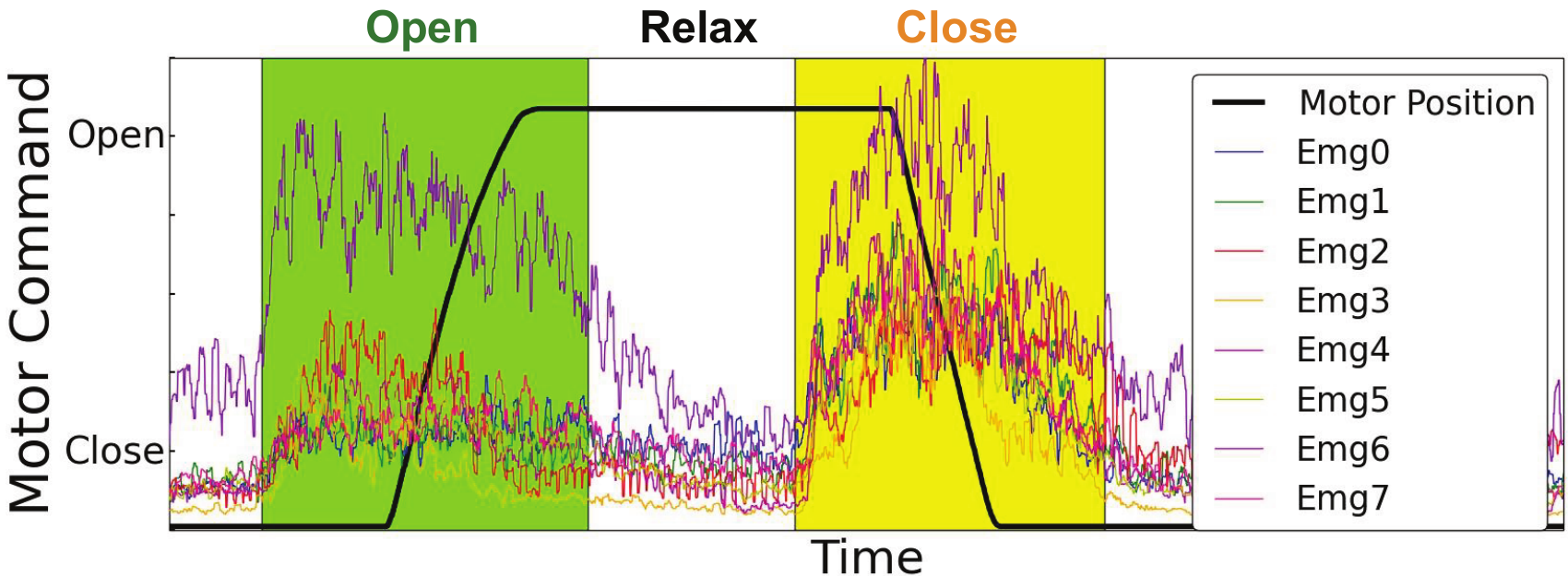}
		\label{fig:emg_training}
	}\\
	\vspace{-0.2cm}
	\subfloat[Bend sensor data and its time derivative]{%
		\includegraphics[width=1.0\linewidth]{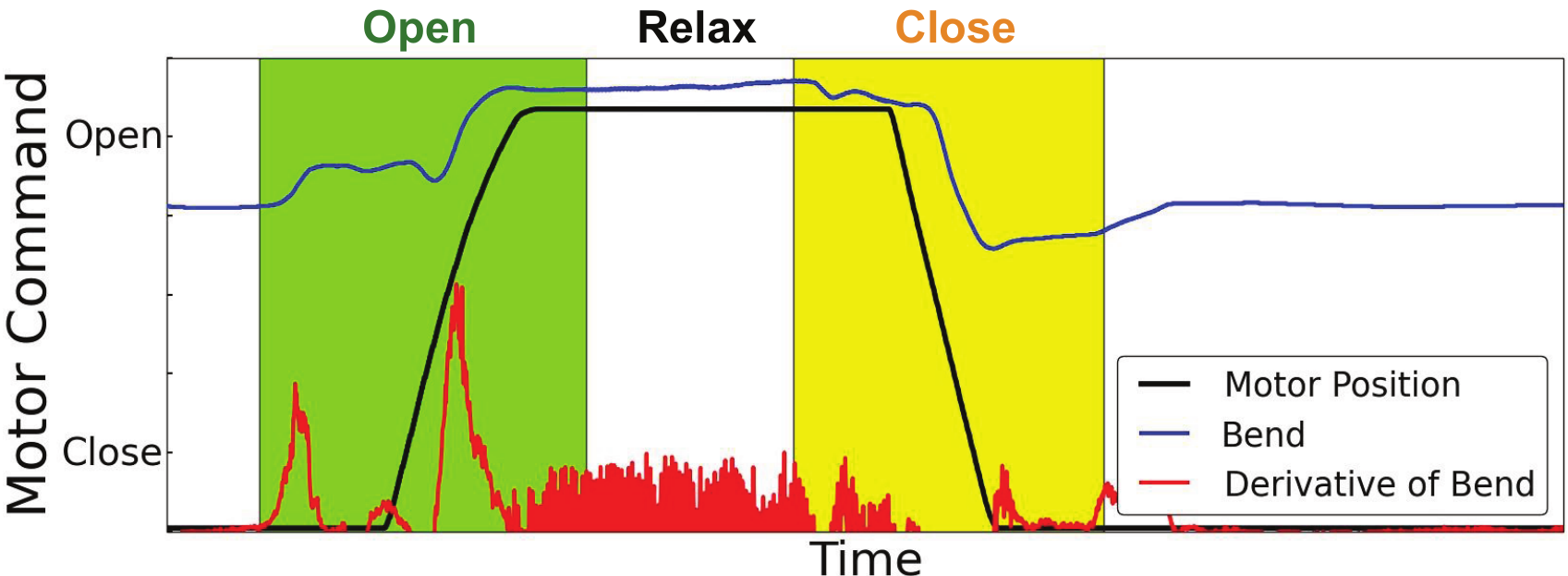}
		\label{fig:bend_training}
	}\\
	\vspace{-0.2cm}
	\subfloat[Fingertip pressure and the time derivative of the pressure]{%
		\includegraphics[width=1.0\linewidth]{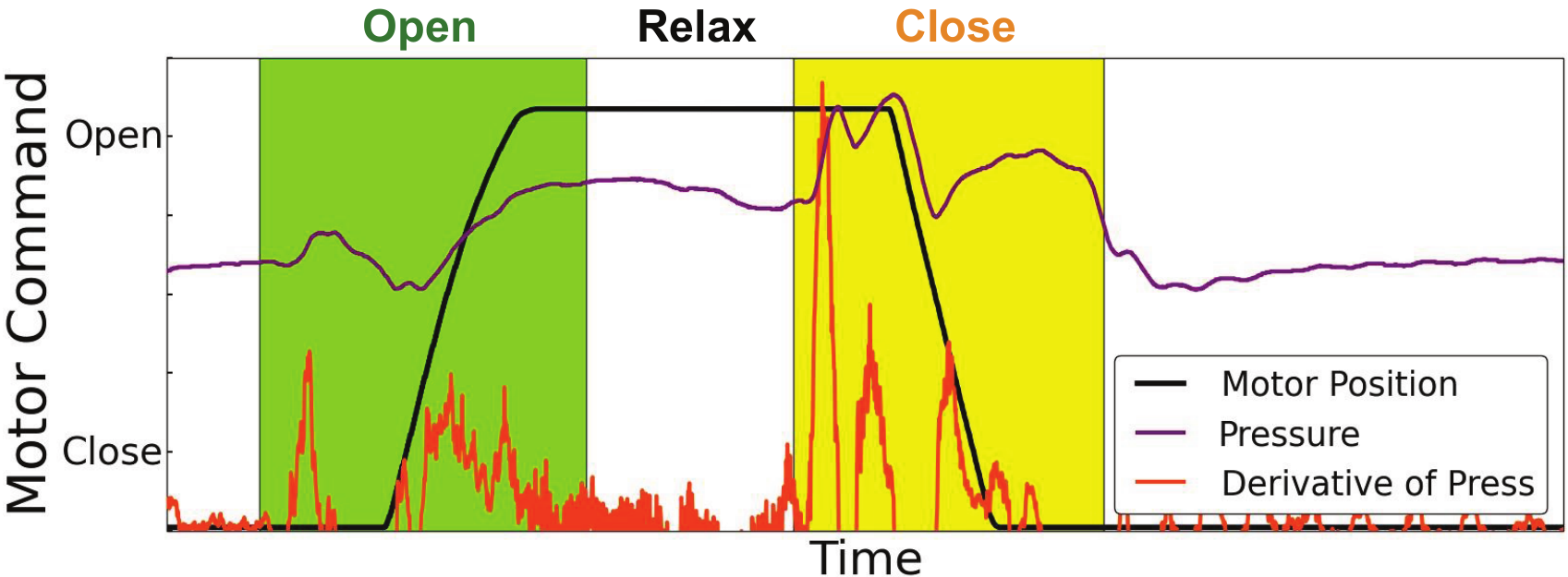}
		\label{fig:press_training}
	}
	\caption{Sensor data recorded while a user with stroke opens (green) and closes (yellow) their affected hand during training. The plots depict the following sequence (verbal instructions are given to the user throughout): the user, with hand at rest, tries to open. After a few seconds, the device is actuated to assist the open. The user relaxes once the hand is open (tendon fully retracted). Then, the user tries to close, despite resistance from the orthosis. Shortly after, the device extends, allowing the hand to close.}
	\vspace{-0.4cm}
\end{figure}

Again, we use the time derivative of the pressure data rather than the
raw data. As shown in Fig.~\ref{fig:press_training}, both the
raw data and the pressure derivative increase soon after the subject
is asked to close their hand, but the derivative provides more robust
cues because the raw data alters over time due to fatigue and
irregular tone.

We fit our exotendon device with pressure sensing using force
sensitive resistors (FSRs). FSRs are compact enough for integration
inside the digit straps which attach the 3D printed fingertip
components to the subject's fingers. Fig.~\ref{fig:sensors}
shows how the FSRs are placed inside the digit straps.

For simplicity, we integrate pressure sensing only on the thumb
because it is the finger which generates the greatest force when the
subject tries to close their hand\corrlab{R1-3c}{~\cite{kargov2004}}. The thumb is also used in all
gross grasping, ensuring we will see interactions between the subject
and any grasped objects.

\section{Controls} \label{sec:controls}

\subsection{EMG Control} \label{subsec:emg_control}

In previous work, we describe an EMG control which uses pattern
recognition to predict user intention~\cite{meeker2017}. Here, we use
the same eight sensor EMG armband (Myo) and a similar pattern
recognition algorithm. Pattern recognition enables the use of
commodity EMG devices, which are easier to don and doff than
medical-grade sensors. Using pattern recognition, the EMG sensors do
not need to be placed on specific muscles in order to identify user
intention.

We place the EMG sensors on the subject's forearm, on the same arm as
the exotendon device. Ipsilateral EMG control harvests the EMG signal
the user makes when they try to open or close their impaired hand,
rather than requiring the user to learn an unrelated motion to control
the device.

The algorithm we use for pattern recognition is described in our
previous work~\cite{meeker2017}. The main difference in this work is
that we aim to predict three possible user intentions rather than two:
to open the hand (\textit{Intent=Open}), to close the hand (\textit{Intent=Closed}), and to relax
(\textit{Intent=Relaxed} - newly introduced here). The addition of the \textit{Intent=Relaxed} class
allows the user to open the hand using the exotendon
device, and then relax their hand while they are positioning their
arm, for example in order to execute a pick and place task, without
having to continue to exert effort to keep the hand open. We believe
this approach can help avoid muscle fatigue.

\corrlab{R1-4a}{To classify user intent at a given time, we input the
  EMG signals collected at that time into a random forest
  classifier. The classifier outputs three values, each being the
  probability that the EMG signals belong to a corresponding intent
  class. These three probabilities are put through a median filter
  (0.5~s window) in order to eliminate spurious predictions.  Finally,
  we compare the output probabilities from the median filter to three
  manually set thresholds. If the probability for a class exceeds the
  threshold, we classify the end result as belonging to that
  class. The end-result belongs to either the \textit{Intent=Open},
  \textit{Intent=Relaxed}, or \textit{Intent=Closed} class. We assign thresholds such
  that only one class can exceed a threshold at a time. If none of the
  thresholds are exceeded, the intent remains the same as at the last
  time step. }

The EMG control can then issue motor commands to the exotendon device
based on the predicted user intention. If the EMG control predicts
that the user's intention is \textit{Intent=Open}, the device is commanded to open
(retract the tendon, thus extending the fingers). If the user's
intention is \textit{Intent=Closed}, the device is commanded to close (extend the
tendon, thus allowing the user to flex the fingers). If the predicted
user intention is \textit{Intent=Relaxed}, we continue to send the previous motor
command to the device.

\subsection{Multimodal Control} \label{subsec:multimodal_control}

We propose two types of multimodal control. Subjects in our target
population display a wide range of impairment patterns. Some cannot
maintain a `close' EMG signal, and others have more voluntary finger
extension. A single sensing modality is limited due to the various
impairment patterns; similarly, multimodal sensing is limited if it
does not fit the subject's impairment pattern.

We propose one kind of multimodal control where bend
sensors detect the user's intention to open the hand, and EMG sensors
detect the user's intent to close the hand. The other multimodal
approach uses pressure sensors to detect the user's intent to close
the hand, and EMG sensors to detect the user's intent to open the
hand. The multimodal approach used for each of our
subjects was chosen based on a qualitative analysis of their abilities,
such as range of voluntary finger extension, and ability to
maintain EMG signals.

\subsubsection{Bend to Open, EMG to Close}
The first multimodal control uses bend sensors to determine when the
exotendon device should open, and EMG sensors determine when the
device should close. Subjects who use this control would typically
have the ability to initiate finger extension, but be unable to
achieve functional extension and have difficulty maintaining an `open'
signal for EMG.

To determine user intent based on voluntary extension, we collect data
from the four bend sensors built into the orthosis. In the current
version, the therapist determines which of the subject's fingers has
the greatest range of voluntary motion and we focus on bend data from
that specific digit; in the future, we plan to integrate the data from
all four sensors. Bend data is then passed through a moving mean
filter with a window size of 0.25~s. We take the derivative
of the resulting signal, which we refer to as $\frac{\partial
  b^i}{\partial t}$ (where the subscript $i$ denotes the digit found
to have the highest voluntary range of motion). Motor commands are
sent as follows:

\begin{mylist}
  \item When the orthosis is in the \textit{Device=Closed} position (tendon extended
    allowing fingers to flex) and $\frac{\partial b^i}{\partial t}$
    exceeds a given threshold $L^B$, the device is commanded to open
    (retract the tendon).
  \item When the orthosis is in the \textit{Device=Open} position (tendon fully
    retracted, or motor stalled) and the EMG classifier predicts
    \textit{Intent=Closed}, the device is commanded to close (extend the tendon).
  \item If neither of the above conditions are met, we continue to send
    the previous motor command to the device.
\end{mylist}

The threshold $L^B$ is determined based on the training data
collected in the procedure described in
Section~\ref{subsec:training}. For the training dataset, we find the
local maxima of $\frac{\partial b^i}{\partial t}$ while we ask the subject
to try to open. We select the smallest value between the
local maxima as $L^B$. If necessary, the experimenter will manually
tune the threshold so the control can enable tasks. After
the threshold is set, it is kept constant throughout all tests
performed by the subject.

When the device is in the \textit{Device=Closed} position, EMG
signals are ignored, as are bend signals when the device is in the \textit{Device=Open}
position. Furthermore, our control will not switch motor commands
while the device is transitioning from \textit{Device=Open} to \textit{Device=Closed} or 
from \textit{Device=Closed} to \textit{Device=Open}. We note that although this 
consideration can reduce rapid oscillations in the motor command, it is 
limiting if the subject only wants to open their hand halfway and then close 
again, for example, when grasping small objects. If the subject starts closing 
their hand before the motor is done transitioning, they will encounter resistance from the orthosis until the transition finishes and the control issues another
command to the motor.

\subsubsection{EMG to Open, Pressure to Close}
For the second kind of multimodal control, EMG sensors determine when
the exotendon device should open and the pressure sensors determine
when the device should close. Subjects who use this control typically
have a clear EMG muscle pattern for `open' and difficulty maintaining
a `close' signal for EMG.

To implement this control, we use data from the thumb pressure sensor. As with
bend data, the raw signal is first passed through a moving average
filter with window size 0.25~s; we then compute the derivative of the
output $\frac{\partial p}{\partial t}$. Motor commands are sent as
follows:
\begin{mylist}
  \item When the orthosis is in the \textit{Device=Closed} position (tendon extended
    allowing fingers to flex) and the EMG classifier predicts \textit{Intent=Open},
    the device is commanded to open (retract the tendon).
  \item When the orthosis is the \textit{Device=Open} position (tendon fully
    retracted, or motor stalled) and $\frac{\partial p}{\partial t}$
    exceeds threshold $L^P$, the device is commanded to close
    (extend the tendon).
  \item If neither of the above conditions are met, we continue to send
    the previous motor command to the device.
\end{mylist}
The threshold $L^P$ is set with a procedure similar to the one
previously described for the bend threshold $L^B$ this time using training data 
while the subject is being asked to try to close. Again, we do not issue 
new commands while the device is transitioning between states.

\subsection{Training with the Exotendon Device}\label{subsec:training}

\remind{R2-2}{Stroke subjects often produce EMG patterns which
	change dramatically depending on arm position, even if the subject's
	intention to open, relax or close the hand remains the same. These EMG
	patterns are further changed by the hand's physical interaction with
	the exotendon device. We therefore train the subjects with their arms
	in different positions and the exotendon device in different states.}

We design our training protocol as follows: the exotendon device
starts in the closed state (tendon is fully extended) and the subject
is asked to relax. Then the subject is asked to try to open their
hand. The experimenter waits three seconds, and as the user continues
to try to open, the experimenter opens the exotendon device (retracts
the tendon) to extend the subject's fingers. The subject continues to
try to open for three seconds after the exotendon device is fully
opened and is then relaxes. Next, the subject is
instructed to close their hand. The experimenter waits three seconds and then
closes the device. The subject continues to try and close for three
seconds after the device has fully closed and then is instructed to relax. 
\corrlab{R1-4c}{During training, subject intent, or ground truth, is given to the program by the experimenter as they simultaneously provide participants with verbal commands.}

The subject repeats the above procedure five times. The first two
times, the subject's arm rests on the table, and the next three times,
the subject raises their arm off the table.

\section{Experiments} \label{experiments}

We evaluate the feasibility of our multimodal controllers when used by subjects with different impairment patterns, using EMG control as a baseline. We selected patients whose EMG patterns showed signs of being abnormal, affected by fatigue and interaction with the orthosis (which, in our experience, is commonplace), but who were still able to complete pick and place tasks using EMG control. Our multimodal control is designed to be robust to different impairments, so we chose subjects with distinct patterns.

\remind{R2-5}{We note that, in this current version of the study, the experimenter plays the important role of selecting the appropriate control mode for a patient. We believe this approach serves to establish the feasibility of multimodal sensing in our context, but is also applicable to real-life scenarios, where an experienced clinician can make similar decisions based on patient observations. Nevertheless, we hope to automate this aspect of the procedure in future work.}

Testing was performed on four chronic subjects with a spasticity level of two or less on the Modified Ashworth Scale (MAS). \corrlab{R1-5a, R2-4}{Subject clinical information can be found in Table~\ref{baseline_info}.} Participants had prior experience with the exotendon device, in varying capacities. Subjects gave informed consent and all testing was approved by the Columbia University Internal Review Board, and performed in a clinical setting under the supervision of an occupational therapist.

\begin{table}[]
	\centering
	\caption{Subject clinical information}
	\label{baseline_info}
	\vspace{-2mm}	
	\begin{tabular}{C{0.85cm}|C{0.83cm}|C{1.0cm}|C{1.4cm}|C{1.83cm}}
		Subject & Gender	& Affected Limb	& Fugl Meyer UE & Box and Blocks 		\\ \hline
		A       & F			& Left			& 26         	& 0     				\\ 
		B       & F			& Right			& 26         	& 0						\\ 
		C       & M			& Right			& 25          	& 6						\\ 
		D       & M			& Right			& 23          	& 0						\\
	\end{tabular}
	\vspace{-5mm}
\end{table}

The experimenter placed the orthosis on the subject's hand and made any necessary sizing adjustments. The subjects were trained using the protocol described in Section~\ref{subsec:training}. 

After training, we asked the subject to perform two types of testing. The first one, designed to isolate the effects of the chosen control method, consists exclusively of performing open-close hand motions. We refer to these as \corrlab{R1-5f}{\textit{Controller Accuracy}} experiments. In these tests, we asked the subject to perform several open and close motions in order to compare the accuracy of the baseline and proposed controls. The experimenter verbally cued the subjects to open and close their hand while providing the program with ground truth for the desired motor command.

The second type of test is designed to verify that the multimodal sensory platform we have developed can be used in a functional context. We refer to these as \textit{Pick and Place} experiments. Here, five blocks (1'' square cubes) were placed in a square pan on a table in front of the subject. The subject was required to start with their hand in a relaxed state, grasp a block, transport it over the median with control and release it onto the tabletop. The task was considered complete when the subject activated the device to extend the digits and released the block. The therapist timed how long it took the subject to pick and place each of the five blocks. For each condition, the subject moved all five blocks three times. \corrlab{R1-5a, R2-4}{Patients were given sufficient time between testing procedures such that order effects which might have been induced by fatigue were negligible.} Each subject was given three minutes of play time to acclimate to each control. 

\corrlab{R1-5g, R2-1, R3-3}{While we designed our pick and place task to minimize the impact of external factors on performance, the nature of functional tasks renders them replete with factors that impact performance. Even such a simple task reflects an individual's shoulder strength, residual fingertip sensation, and grip strength and is not a pure measure of controller efficacy. The number of clinical tests needed to average out the significant effects of all of these compounding factors is beyond the scope of this paper.  We therefore rely on Controller Accuracy to evaluate the proposed controls isolated from other factors, and use Pick and Place experiments simply to illustrate their feasibility in a functional context. For this reason, all subjects completed Controller Accuracy testing, but only Subjects A and B completed Pick and Place testing.}

\remind{R1-5c}{In stroke subjects, fatigue and abnormal coactivation~\cite{dewald1995} can cause EMG patterns to change over time. To study these effects, we also asked the subjects showing most pronounced effects of fatigue and abnormal co-activation (subjects B and D, as observed by the experimenter) to perform all experiments a second time, in a different condition: wearing an arm support system which aids arm movement through gravity compensation. Testing with and without the arm support system helps us evaluate when the multimodal control is most effective.}

During testing, subjects were unaware of the control mode they were using. The controls should be intuitive, so subjects were merely instructed to try to open and close their hand.

\section{Results and Discussion} \label{results}

For Controller Accuracy testing, we compare the output of the EMG and multimodal controls to the ground truth provided by the experimenter. \corrlab{R1-5d, R2-3}{At each time point, the controls can correctly predict an open (true positive), correctly predict a close (true negative), incorrectly predict an open (false positive) or incorrectly predict a close (false negative).} We report the global accuracy, the positive predictive value (PPV), and  the negative predictive value (NPV) for our classifiers~\cite{ortiz2015}. \corrlab{R1-5d}{Global accuracy is the number of true predictions (positive or negative), divided by the number of total predictions. PPV is the number of true positives divided by the number of all positive predictions (whether true or false). NPV is the number of true negatives divided by the number of all negative predictions.}  

\begin{table}[t]
	\centering
	\caption{Results for controller accuracy}
	\vspace{-2mm}
	\label{tab:accuracy_results}  
	\begin{tabular}{C{1.5cm}@{\hspace{1\tabcolsep}}C{1.35cm}|@{\hspace{0.5\tabcolsep}}C{1.2cm}@{\hspace{0.5\tabcolsep}}C{0.8cm}@{\hspace{0.5\tabcolsep}}C{0.8cm}@{\hspace{0.5\tabcolsep}}C{1.0cm}@{\hspace{0.1\tabcolsep}}C{0.7cm}@{\hspace{0.5\tabcolsep}}}
		
		&    	  &		     &	    &       & \multicolumn{2}{c}{Transitions} \\
		Condition & Control Type & Global Accuracy  & PPV  & NPV   & Correct & False
		\\ \hline

		\multirow{2}{1.5cm}{\centering Regular} 	& EMG          
		& 77.9\%	    & 77.1\%	      & 78.8\%	& 4/7			&  \textbf{0.5}		
		\\
		
		& Multimodal   
		& \textbf{83.4\%} & \textbf{81.6\%} & \textbf{84.9\%}	& \textbf{6/7} 	& 1.5 		
		\\ \hline

		\multirow{2}{1.5cm}{\centering With arm support} 	& EMG          
		& 85.2\% 	    & 85.6\%          & 84.8\%        	& 8/9 		& 2 	
		\\
		
		& Multimodal   
		& \textbf{86.0\%} & \textbf{86.6\%} & \textbf{85.6\%} & 8/9  &  \textbf{0}		
		\\
		
	\end{tabular}
	\vspace{-3mm}
\end{table}

Global accuracy can be misleading for EMG pattern recognition controls~\cite{ortiz2015}, so we believe another important metric is the ability to correctly identify transitions between motor commands. A transition is defined as a change in motor command, and a correctly identified transition means a predicted transition which occurs within 1.5 seconds of the ground truth transition. 1.5 seconds allows enough time for the experimenter to give the verbal command and for the subject to react and start performing the motion. (We found that the subjects would often raise their arm off the table before they attempted the instructed hand motion, which increased reaction time.) \corrlab{R1-5e}{The correct transitions are reported with the total number of ground truth transitions. Success for this metric is a number of correctly identified transitions that is close or equal to the total number of ground truth transitions.} We also report the number of false transitions, or transitions which do not have a corresponding ground truth transition. These transitions cause motor oscillations, confusing and frustrating the user. \corrlab{R1-5e}{Success for this metric is a number of false transitions close to zero.}

The results for the Controller Accuracy experiments are shown in Table~\ref{tab:accuracy_results}. We averaged across subjects and show results for experiments performed with arm support (Subjects B and D) and without arm support (all participants). For the Pick and Place testing, we report the time to pick each block and the total time to pick all five blocks, averaged across three trials, and standard error (Table~\ref{tab:pick_and_place_results}). We show results for subjects with arm support (Subject B) and the average result without arm support (Subjects A and B).

\begin{table}[t]
	\centering
	\caption{\corrlab{R2-8}{Results for pick and place tasks}}
	\vspace{-2mm}
	\label{tab:pick_and_place_results}  
	\begin{tabular}{C{1.5cm}@{\hspace{1\tabcolsep}}C{1.35cm}|c@{\hspace{1\tabcolsep}}c}
		
		Condition& Control Type & Each Block & Total  \\ \hline
		\multirow{2}{1.5cm}{\centering Regular} 	& EMG          
		& \textbf{13.4$\pm$1.4} & \textbf{66.8$\pm$5.6}\\
		
		& Multimodal   
		& 15.7$\pm$2.8	& 78.4$\pm$8.4\\ \hline

		\multirow{2}{1.5cm}{\centering With arm support} 	& EMG          
		& 30.1$\pm$6.5	& 150.7$\pm$54.6\\
		
		& Multimodal   
		& \textbf{24.3$\pm$3.8} & \textbf{121.5$\pm$32.4}\\ 
		
	\end{tabular}
	\vspace{-3mm}
\end{table}

In Controller Accuracy testing, in the Regular condition (no arm
support) multimodal control consistently outperformed EMG
control. Both with and without arm support, the global accuracy, PPV,
NPV, and the number of correctly identified transitions within the
given time window were all higher for the multimodal control than for
the EMG control. EMG control only outperformed multimodal control in
the number of false positive transitions predicted.

We believe these results show that the proposed multimodal control can be effective for subjects with a variety of different impairment patterns. Subjects A and D have almost no voluntary extension in their fingers and have difficulty maintaining a `close' EMG signal. On the other hand, Subjects B and C can extend their fingers partially, but have a hard time maintaining an EMG signal for `open'. Our multimodal control can be customized to these different impairment patterns and enables effective orthosis control for both patterns.

With arm support, the accuracy of the two control methods is much
closer. The global accuracy, PPV and NPV are all within 1\% of each
other for the two controls.  We hypothesize that arm support relieves
abnormal muscle coactivation experienced by the subject, and that when
this coactivation diminishes, the subjects have an easier time
maintaining their EMG signals.  These findings tell us two things:
first, they illustrate just how varied post-stroke impairment patterns
can be. The same EMG control had up to a 8.5\% increase in performance
when we started providing arm support. Such a significant change in a
subject-driven control for the same patient underlines the need for
controls which can adapt to a wide range of impairment patterns, both
between patients and as subjects undergo rehabilitation.  Second, they
can tell us where our multimodal control is most useful.  With arm
support, the multimodal control did not help the patient significantly
more than the EMG control. However, it did help subjects more when
they were not provided arm support.  We conclude that our
multimodal control is best suited for patients who experience fatigue
easily and who experience a significant amount of abnormal muscle
coactivation.

\corrlab{R1-5g}{In the pick and place experiments, multimodal
  control was more efficient than EMG control when used with an arm
  support, while EMG control was more efficient with arm
  support. However, the small sample size and the large number of
  additonal factors that affect functional performance (e.g. arm
  strength and control, fingertip sensation, grip strength, chosen
  task strategy, etc.), prevent us from drawing any quantitative
  conclusions. We believe, however, that these results show that a
  multimodal sensory platform can be integrated in a complete
  functional task, highlighted by the fact that all participants
  completed the task using both control mechanisms, despite having `no
  to poor' upper extremity capacity (as defined in \cite{singer2017}).}

\section{Conclusion and Future Work}

In this paper, we incorporate EMG, bend, and pressure sensors into an
exotendon framework to create a multimodal sensing and interaction
platform for a hand orthosis. We believe the future of robust controls
for orthoses involves multiple sensing modalities which complement
each other to inform controls. The bend and pressure sensors give us
information about user intent if subjects display certain impairment
patterns we have observed in many stroke patients.

We propose two multimodal control modes, tailored to the
different impairment patterns we have observed. Controls that can cope
with many impairment patterns are necessary because these patterns
vary across subjects; one patient could even display several patterns
as they undergo therapy post-stroke. This is a preliminary study with
a limited sample size; however, our results show that multimodal
controls can be adapted to different impairment patterns and can help functional tasks. This is a first step
towards the development of robust, flexible controls, which could play
an important part in deploying robotic rehabilitation to a large
population of stroke patients.

In the future, we would like to add more sensing modalities to our
device, such as bend sensors on the MCP joints and IMU sensors to
provide information about finger positions. IMUs and bend sensors have
different limitations and strengths and will complement each other to
jointly characterize finger movement. We also believe that a
multimodal sensing platform, like the one developed here, could use
its sensors not only for control, but also to track subject progress
and rehabilitation. This will enhance our understanding of phenomena
such as muscle spasticity and abnormal muscle synergies. It could also
help us understand how impairment patterns develop over time as
patients undergo rehabilitation.

\remind{R3-2}{To expand our study of robust multimodal controls, in the future, we
would like to develop a control which uses inputs from all sensor
types simultaneously.} We believe that
the future of these multimodal controls lies in the sensors' ability
to complement and augment each other, and that such a control can
continuously adapt and learn to predict user intent. 
The multimodal control predicted
signals that the EMG-only classifier missed (i.e. the correctly predicted transitions 
in Table~\ref{tab:accuracy_results}). This suggests that our
multimodal control could be used to continue training the EMG
classifier during real-time operations. Such continuous adaptation
will also play an important role as the field transitions from
controlled sessions to in-home environments. \balance

\bibliographystyle{IEEEtran}
\bibliography{bib/orthoses,bib/stroke}

\begin{thebibliography}{10}
\providecommand{\url}[1]{#1}
\csname url@rmstyle\endcsname
\providecommand{\newblock}{\relax}
\providecommand{\bibinfo}[2]{#2}
\providecommand\BIBentrySTDinterwordspacing{\spaceskip=0pt\relax}
\providecommand\BIBentryALTinterwordstretchfactor{4}
\providecommand\BIBentryALTinterwordspacing{\spaceskip=\fontdimen2\font plus
\BIBentryALTinterwordstretchfactor\fontdimen3\font minus
  \fontdimen4\font\relax}
\providecommand\BIBforeignlanguage[2]{{%
\expandafter\ifx\csname l@#1\endcsname\relax
\typeout{** WARNING: IEEEtran.bst: No hyphenation pattern has been}%
\typeout{** loaded for the language `#1'. Using the pattern for}%
\typeout{** the default language instead.}%
\else
\language=\csname l@#1\endcsname
\fi
#2}}

\bibitem{park2016}
S.~Park, L.~Bishop, T.~Post, Y.~Xiao, J.~Stein, and M.~Ciocarlie, ``On the
  feasibility of wearable exotendon networks for whole-hand movement patterns
  in stroke patients,'' in \emph{Robotics and Automation (ICRA), 2016 IEEE
  Intl. Conf. on}.\hskip 1em plus 0.5em minus 0.4em\relax IEEE, 2016, pp.
  3729--3735.

\bibitem{biggar2016}
S.~Biggar and W.~Yao, ``Design and evaluation of a soft and wearable robotic
  glove for hand rehabilitation,'' \emph{IEEE Transactions on Neural Systems
  and Rehab. Engineering}, vol.~24, pp. 1071--1080, 2016.

\bibitem{in2015}
H.~In, B.~B. Kang, M.~Sin, and K.-J. Cho, ``Exo-glove: a wearable robot for the
  hand with a soft tendon routing system,'' \emph{IEEE Robot Autom Mag},
  vol.~22, no.~1, 2015.

\bibitem{kang2016}
B.~B. Kang, H.~Lee, H.~In, U.~Jeong, J.~Chung, and K.-J. Cho, ``Development of
  a polymer-based tendon-driven wearable robotic hand,'' in \emph{IEEE Intl.
  Conf. on Robotics and Automation}, 2016.

\bibitem{zhao2016}
H.~Zhao, J.~Jalving, R.~Huang, R.~Knepper, A.~Ruina, and R.~Shepherd, ``A
  helping hand: Soft orthosis with integrated optical strain sensors and emg
  control,'' \emph{IEEE Robot Autom Mag}, vol.~23, no.~3, 2016.

\bibitem{polygerinos2015}
P.~Polygerinos, K.~C. Galloway, S.~Sanan, M.~Herman, and C.~J. Walsh, ``Emg
  controlled soft robotic glove for assistance during activities of daily
  living,'' in \emph{IEEE Intl. Conf. on Rehabilitation Robotics}, 2015.

\bibitem{meeker2017}
C.~Meeker, S.~Park, L.~Bishop, J.~Stein, and M.~Ciocarlie, ``Emg pattern
  classification to control a hand orthosis for functional grasp assistance
  after stroke,'' in \emph{IEEE Intl. Conf. on Rehabilitation Robotics}, 2017.

\bibitem{cesqui2013}
B.~Cesqui, P.~Tropea, S.~Micera, and H.~I. Krebs, ``Emg-based pattern
  recognition approach in post stroke robot-aided rehabilitation: a feasibility
  study,'' \emph{Journal of neuroengineering and rehabilitation}, vol.~10,
  no.~1, p.~75, 2013.

\bibitem{ochoa2009}
J.~M. Ochoa, Y.~J.~D. Narasimhan, and D.~G. Kamper, ``Development of a portable
  actuated orthotic glove to facilitate gross extension of the digits for
  therapeutic training after stroke,'' in \emph{Engineering in Medicine and
  Biology Society, 2009. EMBC 2009. Annual Intl. Conf. of the IEEE}.\hskip 1em
  plus 0.5em minus 0.4em\relax IEEE, 2009, pp. 6918--6921.

\bibitem{chu2018}
C.-Y. Chu and R.~M. Patterson, ``Soft robotic devices for hand rehabilitation
  and assistance: a narrative review,'' \emph{J Neuroeng Rehabil}, vol.~15,
  no.~1, 2018.

\bibitem{hu2013}
X.~Hu, K.~Tong, X.~Wei, W.~Rong, E.~Susanto, and S.~Ho, ``The effects of
  post-stroke upper-limb training with an electromyography (emg)-driven hand
  robot,'' \emph{Journal of Electromyography and Kinesiology}, vol.~23, no.~5,
  pp. 1065--1074, 2013.

\bibitem{lu2018}
Z.~Lu, R.~K.-y. Tong, X.~Zhang, S.~Li, and P.~Zhou, ``Myoelectric pattern
  recognition for controlling a robotic hand: A feasibility study in stroke,''
  \emph{IEEE Transactions on Biomedical Engineering}, 2018.

\bibitem{lucas2004}
L.~Lucas, M.~DiCicco, and Y.~Matsuoka, ``An emg-controlled hand exoskeleton for
  natural pinching,'' \emph{J. of Robot. and Mechatronics}, vol.~16, pp.
  482--488, 2004.

\bibitem{hussain2017}
I.~Hussain, G.~Spagnoletti, G.~Salvietti, and D.~Prattichizzo, ``Toward
  wearable supernumerary robotic fingers to compensate missing grasping
  abilities in hemiparetic upper limb,'' \emph{The Intl. Journal of Robotics
  Research}, vol.~36, no. 13-14, pp. 1414--1436, 2017.

\bibitem{thielbar2017}
K.~O. Thielbar, K.~M. Triandafilou, H.~C. Fischer, J.~M. O’Toole, M.~L.
  Corrigan, J.~M. Ochoa, M.~E. Stoykov, and D.~G. Kamper, ``Benefits of using a
  voice and emg-driven actuated glove to support occupational therapy for
  stroke survivors,'' \emph{IEEE Transactions on Neural Systems and
  Rehabilitation Engineering}, vol.~25, no.~3, pp. 297--305, 2017.

\bibitem{yap2016a}
H.~K. Yap, B.~W. Ang, J.~H. Lim, J.~C. Goh, and C.-H. Yeow, ``A
  fabric-regulated soft robotic glove with user intent detection using emg and
  rfid for hand assistive application,'' in \emph{IEEE Int. Conf. on Robotics
  and Automation}, 2016.

\bibitem{guo2017}
W.~Guo, X.~Sheng, H.~Liu, and X.~Zhu, ``Mechanomyography assisted myoeletric
  sensing for upper-extremity prostheses: A hybrid approach,'' \emph{IEEE
  Sensors Journal}, vol.~17, pp. 3100--3108, 2017.

\bibitem{wolczowski2017}
A.~Wo{\l}czowski and R.~Zdunek, ``Electromyography and mechanomyography signal
  recognition: experimental analysis using multi-way array decomposition
  methods,'' \emph{Biocybern Biomed Eng}, vol.~37, no.~1, 2017.

\bibitem{ibitoye2014}
M.~O. Ibitoye, N.~A. Hamzaid, J.~M. Zuniga, and A.~K.~A. Wahab,
  ``Mechanomyography and muscle function assessment: A review of current state
  and prospects,'' \emph{Clin Biomech}, vol.~29, no.~6, 2014.

\bibitem{luo2005}
X.~Luo, T.~Kline, H.~C. Fischer, K.~A. Stubblefield, R.~V. Kenyon, and D.~G.
  Kamper, ``Integration of augmented reality and assistive devices for
  post-stroke hand opening rehabilitation,'' in \emph{IEEE-EMBS Intl. Conf. on
  Engineering in Medicine and Biology Society}, 2005.

\bibitem{yap2016b}
H.~K. Yap, A.~Mao, J.~C. Goh, and C.-H. Yeow, ``Design of a wearable fmg
  sensing system for user intent detection during hand rehabilitation with a
  soft robotic glove,'' in \emph{IEEE Intl. Conf. on Biomedical Robotics and
  Biomechatronics}, 2016.

\bibitem{nilsson2012}
M.~Nilsson, J.~Ingvast, J.~Wikander, and H.~von Holst, ``The soft extra muscle
  system for improving the grasping capability in neurological
  rehabilitation,'' in \emph{IEEE-EMBS Conf. on Biomed Eng Sci}, 2012.

\bibitem{steinkamp2014}
J.~M. Steinkamp, L.~J. Brattain, C.~J. Walsh, and R.~D. Howe, ``Point cloud
  context analysis for rehabilitation grasping assistance.''

\bibitem{ates2017}
S.~Ates, C.~J. Haarman, and A.~H. Stienen, ``Script passive orthosis: design of
  interactive hand and wrist exoskeleton for rehabilitation at home after
  stroke,'' \emph{Autonomous Robots}, vol.~41, pp. 711--723, 2017.

\bibitem{amirabdollahian2014}
F.~Amirabdollahian, S.~Ates, A.~Basteris, A.~Cesario, J.~Buurke, H.~Hermens,
  D.~Hofs, E.~Johansson, G.~Mountain, N.~Nasr, \emph{et~al.}, ``Design,
  development and deployment of a hand/wrist exoskeleton for home-based
  rehabilitation after stroke-script project,'' \emph{Robotica}, vol.~32,
  no.~8, pp. 1331--1346, 2014.

\bibitem{suzuki2007}
K.~Suzuki, G.~Mito, H.~Kawamoto, Y.~Hasegawa, and Y.~Sankai, ``Intention-based
  walking support for paraplegia patients with robot suit hal,'' \emph{Advanced
  Robotics}, vol.~21, no.~12, pp. 1441--1469, 2007.

\bibitem{villa2015}
A.~Villa-Parra, D.~Delisle-Rodr{\'\i}guez, A.~L{\'o}pez-Delis, T.~Bastos-Filho,
  R.~Sagar{\'o}, and A.~Frizera-Neto, ``Towards a robotic knee exoskeleton
  control based on human motion intention through eeg and semgsignals,''
  \emph{Proced. Manuf}, vol.~3, pp. 1379--1386, 2015.

\bibitem{park2018}
S.~Park, L.~Weber, L.~Bishop, J.~Stein, and M.~Ciocarlie, ``Design and
  development of effective transmission mechanisms on a tendon driven hand
  orthosis for stroke patients,'' in \emph{Robotics and Automation (ICRA), 2018
  IEEE Intl. Conf. on}.\hskip 1em plus 0.5em minus 0.4em\relax IEEE, 2018, pp.
  2281--2287.

\bibitem{arata2013}
J.~Arata, K.~Ohmoto, R.~Gassert, O.~Lambercy, H.~Fujimoto, and I.~Wada, ``A new
  hand exoskeleton device for rehabilitation using a three-layered sliding
  spring mechanism,'' in \emph{Robotics and Automation (ICRA), 2013 IEEE Intl.
  Conf. on}.\hskip 1em plus 0.5em minus 0.4em\relax IEEE, 2013, pp. 3902--3907.

\bibitem{kamper2006}
D.~G. Kamper, H.~C. Fischer, E.~G. Cruz, and W.~Z. Rymer, ``Weakness is the
  primary contributor to finger impairment in chronic stroke,'' \emph{Archives
  physical medicine and rehab.}, vol.~87, pp. 1262--1269, 2006.

\bibitem{kargov2004}
A.~Kargov, C.~Pylatiuk, J.~Martin, S.~Schulz, and L.~D{\"o}derlein, ``A
  comparison of the grip force distribution in natural hands and in prosthetic
  hands,'' \emph{Disability and Rehab.}, vol.~26, no.~12, 2004.

\bibitem{dewald1995}
J.~P. Dewald, P.~S. Pope, J.~D. Given, T.~S. Buchanan, and W.~Z. Rymer,
  ``Abnormal muscle coactivation patterns during isometric torque generation at
  the elbow and shoulder in hemiparetic subjects,'' \emph{Brain}, vol. 118,
  no.~2, pp. 495--510, 1995.

\bibitem{ortiz2015}
M.~Ortiz-Catalan, F.~Rouhani, R.~Br{\aa}nemark, and B.~H{\aa}kansson, ``Offline
  accuracy: a potentially misleading metric in myoelectric pattern recognition
  for prosthetic control,'' in \emph{IEEE Eng Med Biol Soc}.\hskip 1em plus
  0.5em minus 0.4em\relax IEEE, 2015.

\bibitem{singer2017}
B.~Singer and J.~Garcia-Vega, ``The fugl-meyer upper extremity scale,'' \emph{J
  Physiother}, vol.~63, no.~1, p.~53, 2017.

\end{thebibliography}
\end{document}